\documentclass[11pt,letterpaper]{article}
\usepackage{arxiv}
\usepackage{fontspec}
\usepackage{amsmath,amssymb}
\usepackage{unicode-math}
\usepackage{microtype}
\usepackage{graphicx,adjustbox}
\usepackage{longtable,booktabs,array,calc}
\usepackage{caption}
\usepackage[authoryear,round]{natbib}
\setcitestyle{aysep={,},citesep={;}}
\usepackage{xcolor}
\definecolor{linkblue}{HTML}{315E89}
\usepackage{xurl}
\usepackage{hyperref}
\usepackage{bookmark}
\hypersetup{colorlinks=true,allcolors=linkblue,
  pdftitle={Whom to Trust About the
Unseen},pdfauthor={Ruijia Guan},pdflang={en}}
\renewcommand{\normalsize}{\fontsize{11}{13.5}\selectfont}
\normalsize
\renewcommand{\headeright}{Preprint}
\renewcommand{\undertitle}{}
\renewcommand{\shorttitle}{\small Whom to Trust About the Unseen}
\fancypagestyle{plain}{%
  \fancyhf{}\chead{\shorttitle}\rhead{\scshape\footnotesize\headeright}
  \cfoot{\thepage}}
\providecommand{\tightlist}{\setlength{\itemsep}{0pt}\setlength{\parskip}{0pt}}
\newcommand{\pandocbounded}[1]{%
  \begin{adjustbox}{max width=\linewidth,max totalheight=.82\textheight}#1\end{adjustbox}}
\title{Whom to Trust About the Unseen\\[5pt]{\large\normalfont How
Language Models Weigh Spiritual and Scientific Testimony}}
\author{Ruijia Guan\\
Symbolic Systems Program, Stanford University}
\date{}
\begin{document}
\maketitle
\begin{abstract}

Much of what people believe about the unseen, from germs to spirits,
rests on testimony they cannot check. We find that language models do
not weigh such testimony by a single standard. Across three studies and
six models from five families, models treat the spiritual as an
epistemic domain of its own, with its own credible sources, its own
standard for an adequate answer, and its own limits on whose word
carries in. They credit a New-Age statement more from a spiritual leader
than from a scientist, the reverse of people, and turn to the scientist
only once the statement borrows the language of physics. When a trusted
expert calls a belief's mechanism a mystery, they lose confidence in
science but not in religion, while people lose confidence in both. And a
record of accurate forecasts makes them trust a source's natural
explanations but barely its spiritual ones, and trust even the natural
ones less once the source credits a spirit. For most models this
boundary holds even where spirits are assumed to exist. Language models
thus draw a sharper line between spiritual and natural knowledge than
people do.

\end{abstract}

\section{Introduction}\label{introduction}

No one verifies firsthand that germs cause fevers, or that the dead
linger near a burial ground. Beliefs about unseen entities, whether
natural or spiritual, rest largely on other people's testimony, and
deciding what to accept depends on whom we trust and about what
\citep{harris2006}. Language models are now among the sources people
routinely consult when weighing claims no one in the conversation can
directly check. This paper asks whether models evaluate spiritual
testimony by the same epistemic rules they apply to natural testimony.

Evaluating any piece of testimony raises three basic questions, and
human research shows that people treat spiritual and natural claims
differently on each one. First, \emph{who counts as a credible source?}
Across 24 countries, adults judge the same obscure statement more
credible when attributed to a scientist than to a spiritual guru
\citep{hoogeveen2022}. Second, \emph{what counts as an adequate
explanation?} When a trusted expert calls an unexplained mechanism ``a
mystery,'' people lose more confidence in a scientific belief than in a
religious one \citep{davoodi2022}. Third, \emph{how far does a track
record of accuracy extend?} Even young children prefer informants who
have been accurate in the past \citep{koenig2004, pasquini2007}, yet
they restrict that trust by domain and by how the informant achieved
accuracy \citep{koenig2011, einav2011}.

Prior studies of epistemic trust in language models have examined source
preferences and accuracy records without directly comparing spiritual
and natural claims. Models display systematic biases toward particular
source labels \citep{khan2026} and can condition trust on an informant's
interaction history \citep{zhou2026}, while the question of how source
credibility should depend on the claim itself has been flagged as an
open problem \citep{ashkinaze2026}.

We investigate all three questions in the same six language models,
drawn from five model families. Study 1 adapts the scientist-versus-guru
paradigm \citep{hoogeveen2022} and Study 2 adapts the expert-ignorance
paradigm \citep{davoodi2022}, benchmarking model judgments directly
against the original human data. Study 3 introduces a new paradigm set
in a fictional river village, where two informants build unequal track
records on verifiable river forecasts before endorsing matched natural
and spiritual explanations of the same local events.

Taken together, the three studies point to a single conclusion. Language
models partition the spiritual into an epistemic domain of its own,
drawing a sharper boundary between spiritual and natural knowledge than
people do. First, credibility in these models tracks role--content fit
rather than general scientific deference. Whereas humans favor the
scientist on obscure New-Age statements, models favor the spiritual
leader and switch to the scientist only when the same empty syntax is
rewritten in physics vocabulary. Second, models apply stricter
explanatory norms to science than to religion. Calling a belief's
mechanism ``a mystery'' lowers humans' confidence in both domains, yet
leaves models' religious confidence virtually untouched while eroding
their scientific confidence. Third, an earned empirical record stops at
the spiritual boundary. Five of the six models transfer a forecaster's
accuracy to hidden natural causes while granting almost none of that
credit to spiritual causes, and they penalize the forecaster even on
natural claims once the forecaster credits a river spirit. When we
reframe Study 3 inside a village where spirits are assumed to exist,
lifting spiritual claims well off the bottom of the scale, this gap
persists in four of those five models and closes only in the fifth.

\section{Related work}\label{related-work}

\subsection{Testimony and the scope of
expertise}\label{testimony-and-the-scope-of-expertise}

Developmental psychology shows that children learn much of what they
know from testimony while remaining remarkably selective about whom to
trust. Preschoolers track which informants have labeled familiar objects
accurately and prefer them for novel words \citep{koenig2004}, and
four-year-olds monitor relative accuracy across speakers
\citep{pasquini2007}. Crucially, this trust is neither blind nor
unlimited. Children favor informants who were accurate on their own over
those who relied on help \citep{einav2011}, expect a specialist's
knowledge to stop at domain boundaries \citep{landrum2015}, and weigh an
informant's credentials against a claim's prior plausibility
\citep{lane2015}. Moreover, domain expertise generalizes narrowly
whereas incompetence tends to generalize broadly \citep{koenig2011}, and
children's confidence in unseen entities depends on both the testimony
they hear and the kind of entity in question \citep{harris2006}. Study 3
builds directly on this developmental framework by separating a source's
verified success on one task from the reach of that credibility onto new
domains.

Two adult studies provide our human benchmarks for the first two
questions. Hoogeveen and colleagues document the ``Einstein effect,'' in
which attributing an obscure statement to a scientist rather than a
spiritual guru boosts perceived credibility across cultures
\citep{hoogeveen2022}. Davoodi and Lombrozo show that the same admission
of expert ignorance affects scientific and religious beliefs
asymmetrically, with ``mystery'' carrying far more epistemic threat in
science than in religion \citep{davoodi2022}.

\subsection{Source trust in language
models}\label{source-trust-in-language-models}

Recent work has begun mapping how language models weigh source identity
and reliability. Khan and colleagues show that models hold systematic
preferences among attributed sources and respond strongly to framing
\citep{khan2026}; Study 1 builds on these label effects by testing how
their direction reverses with statement content. Zhou and colleagues
show through Epistemic Context Learning that models can calibrate trust
from peers' interaction histories \citep{zhou2026}; Study 3 likewise
provides an accuracy record in context, but systematically varies both
the target claim's domain and how the informant explains their own
competence.

Our closest benchmark is Learn2Discern \citep{ashkinaze2026}, which
pairs perturbed answers with real news outlets of externally scored
reliability. That work distinguishes misestimating a source's
reliability from failing to use a correct estimate, and highlights
claim-dependent credibility as a core limitation of global source
ratings. Study 3 isolates that exact problem. By assigning controlled
records to fictional informants, holding each target claim identical
across source conditions, and pairing natural and spiritual explanations
of the same event, we remove real-world source priors. Because neither
causal explanation has a ground-truth answer within the task, a smaller
reliability effect on spiritual claims measures the scope of epistemic
transfer rather than factual error.

\section{Models and response formats}\label{models-and-response-formats}

We evaluate six language models spanning five model families (Claude,
GPT, DeepSeek, Qwen, and Llama). Specifically, we test Claude Haiku 4.5
({\small\ttfamily\nolinkurl{claude-haiku-4-5}}), Claude Sonnet 4.5
({\small\ttfamily\nolinkurl{claude-sonnet-4-5}}), GPT-5
({\small\ttfamily\nolinkurl{gpt-5}}), DeepSeek-V3-0324
({\small\ttfamily\nolinkurl{deepseek/deepseek-chat-v3-0324}}),
Qwen3.5-27B ({\small\ttfamily\nolinkurl{qwen/qwen3.5-27b}}), and
Llama-3.3-70B-Instruct
({\small\ttfamily\nolinkurl{meta-llama/llama-3.3-70b-instruct}}). The
Anthropic and OpenAI models were queried through their official APIs,
while the remaining three were queried through OpenRouter with reasoning
disabled; Qwen was pinned to a single backend (Novita) with fallback
routing turned off. Sampling temperature was set to 1 for all Anthropic
and OpenRouter calls, whereas GPT-5 was run with minimal reasoning
effort (which does not accept a temperature parameter).

Studies 1 and 2 compare two response formats. In the
\emph{participant-conditioned} format, we render an actual human
participant's demographic and attitude survey responses as sentences in
the system prompt and instruct the model to answer as that person,
linking each model trial directly to a human trial. In the \emph{direct}
format, the model answers as itself with no persona prompt. Study 3 uses
the direct format (plus a village-resident frame described in
Section~\ref{design}) and allows models to reason briefly before stating
a final numeric answer. Full details on data handling, parsing, and
secondary-condition coverage appear in
Appendix~\ref{appendix-a.-data-handling-coverage-and-supplementary-results}.

\section{Study 1: Who sounds
credible?}\label{study-1-who-sounds-credible}

Study 1 examines the first question of epistemic trust, namely how
source identity shapes credibility across spiritual and scientific
registers.

\subsection{Method}\label{method}

We adapted the Einstein-effect paradigm \citep{hoogeveen2022}, which
attributes the same obscure, pseudo-profound New-Age statements (for
example, about ``exploring the cosmos itself as an interface between
faith and empathy'') either to a particle physicist or to a spiritual
leader. From the original dataset of 10,195 participants across 24
countries, we sampled 301 profiles stratified by country and religiosity
tercile. Each model completed the exact two trials seen by each sampled
participant, with the same statement--source pairings, rating
credibility from 1 (not at all credible) to 7 (extremely credible) in a
separate call per rating.

To test whether source preferences depend on the register of the
statement, we wrote a matched set of two physics-worded statements that
preserved the original sentence frames while replacing New-Age terms
with technical vocabulary (such as ``entropy,'' ``symmetry,'' and
``decoherence''). Like the originals, these statements are intentionally
meaningless. Both content sets were run in both the
participant-conditioned and direct formats, yielding twelve
model-by-format cells in all. Human reference data exist only for the
original statements; all four statements and prompt templates appear in
Appendix~\ref{appendix-b.-stimuli-and-prompts}.

\subsection{Results}\label{results}

\begin{figure}
\centering
\pandocbounded{\includegraphics[keepaspectratio,alt={Study 1: mean credibility by source and statement content, participant-conditioned format, with the human sample at left (original statements only). Error bars are standard errors over the 301 participant-linked trials.}]{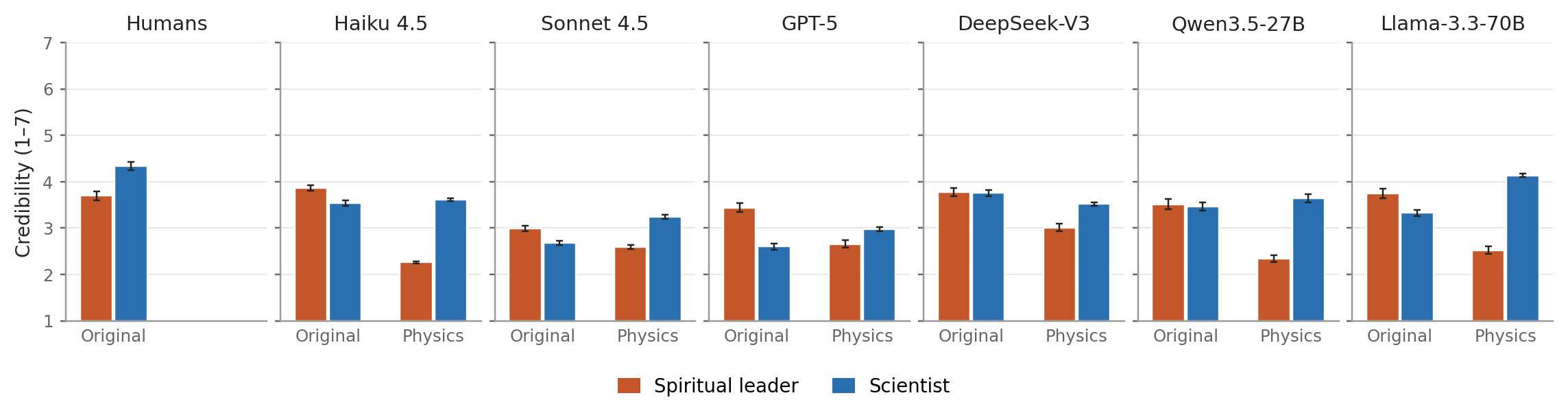}}
\caption{Study 1: mean credibility by source and statement content,
participant-conditioned format, with the human sample at left (original
statements only). Error bars are standard errors over the 301
participant-linked trials.}\label{fig:study1-levels}
\end{figure}

In the human sample, the scientist outscores the spiritual leader on the
original statements (4.34 vs.~3.69, a gap of +0.64;
Figure~\ref{fig:study1-levels}). Language models reverse this
preference. Across the twelve model-by-format cells, ten rate the
spiritual leader as \emph{more} credible than the scientist on the
original statements (Table~\ref{tab:1}). In the remaining two cells
(participant-conditioned DeepSeek and Qwen), the two sources are rated
indistinguishably (−0.02 and −0.05, both within one standard error of
zero).

Rewriting the statements in physics vocabulary flips the preference back
toward the scientist, largely by undercutting the spiritual leader.
Across all twelve cells, the spiritual leader's rating drops
substantially (by 0.40 to 4.60 points), whereas the scientist's rating
rises more modestly (increasing in ten cells, by at most 0.89 points).
After rewording, ten of the twelve cells favor the scientist; the two
exceptions are direct-format Sonnet, which ties the two sources (0.00),
and direct-format Llama, which still leans slightly toward the spiritual
leader (−0.46).

\begin{longtable}[]{@{}lrrrr@{}}
\caption{Scientist minus spiritual-leader credibility on the seven-point
scale. P denotes participant-conditioned and D direct responses. The
human sample's contrast on the original statements is +0.64.
Physics-worded statements have no human
comparison.}\label{tab:1}\tabularnewline
\toprule\noalign{}
Model & Original P & Physics P & Original D & Physics D \\
\midrule\noalign{}
\endfirsthead
\toprule\noalign{}
Model & Original P & Physics P & Original D & Physics D \\
\midrule\noalign{}
\endhead
\bottomrule\noalign{}
\endlastfoot
Haiku 4.5 & -0.33 & +1.36 & -1.02 & +0.97 \\
Sonnet 4.5 & -0.32 & +0.64 & -0.53 & +0.00 \\
GPT-5 & -0.84 & +0.33 & -0.92 & +0.36 \\
DeepSeek-V3 & -0.02 & +0.50 & -1.09 & +1.02 \\
Qwen3.5-27B & -0.05 & +1.30 & -4.60 & +0.53 \\
Llama-3.3-70B & -0.42 & +1.61 & -3.92 & -0.46 \\
\end{longtable}

Because the direct format omits participant profiles, each content set
reduces to four prompts repeated across 301 trials, and models show very
little trial-to-trial variance on repeated prompts
(Appendix~\ref{appendix-a.-data-handling-coverage-and-supplementary-results},
Figure~\ref{fig:study1-levels-direct}). This rigidity explains the two
largest contrasts in Table~\ref{tab:1}, Qwen-direct (−4.60) and
Llama-direct (−3.92), where both models assign the scientist's original
statements the scale minimum of 1 on every trial while rating the
spiritual leader between 4 and 7.

Overall, source credibility in these models reflects role--content fit
rather than a general preference for scientific authority. Whereas
humans deferred to the scientist even on New-Age prose, models treat a
spiritual leader as the right voice for spiritual-sounding language and
penalize him sharply when he borrows the vocabulary of physics, even
though the physics statements are just as empty as the originals. With
only two statements per content set, Study 1 establishes that this
reversal occurs without mapping the full range of phrasings that trigger
it.

\section{Study 2: What counts as an adequate
answer?}\label{study-2-what-counts-as-an-adequate-answer}

Study 2 turns to the second question of epistemic trust, testing whether
an expert's admission of ignorance carries different consequences for
religious and scientific beliefs.

\subsection{Method}\label{method-1}

We adapted Studies 2 and 3 of Davoodi and Lombrozo \citep{davoodi2022}.
In Part A, each trial focuses on a scientific or religious belief held
by an original human participant who did not know its underlying
mechanism. In three sequential turns on a 1--7 scale, the model reports
initial confidence in the belief, estimates how likely experts are to
understand its mechanism, and then projects its confidence after a
trusted expert who shares the belief describes the mechanism as either
``It's unknown'' or ``It's a mystery.'' Following the original inclusion
filters, Part A uses 1,437 participant-by-domain observations from 1,004
participants. Each participant's religious and scientific blocks appear
in the same conversation in a seeded random order. To verify that
confidence changes reflect the expert's response rather than mere
repetition, we also ran a no-expert control on Haiku, Sonnet, and GPT-5
that simply asked for confidence a second time.

Part B probes how models interpret the expert's answer across 1,562
observations from 974 participants, using the first of five fixed
scientific or religious beliefs each participant endorsed without
knowing its mechanism. Models rate eleven statements on a −3 to +3
agreement scale, forming an \emph{epistemic composite} of three items
rating the expert's truth-seeking, knowledge, and curiosity, and a
\emph{threat composite} of eight items measuring how much the expert's
reply threatens or devalues the belief and its broader domain. Part B
was run on Haiku, Sonnet, and GPT-5.

Our primary summary for Part A is the interaction between domain
(science = 1, religion = 0) and wording (``mystery'' = 1, ``unknown'' =
0) in a mixed-effects model of confidence change with a participant
random intercept. A negative interaction indicates that calling a
mechanism ``a mystery'' rather than ``unknown'' costs scientific beliefs
more confidence than religious beliefs.

\subsection{Results}\label{results-1}

\begin{figure}
\centering
\pandocbounded{\includegraphics[keepaspectratio,alt={Study 2: confidence before and after the expert's answer, participant-conditioned format, with the human data at left. Initial confidence is pooled over the two wording groups, which are assigned after it is measured. Error bars are standard errors over participant-by-domain rows.}]{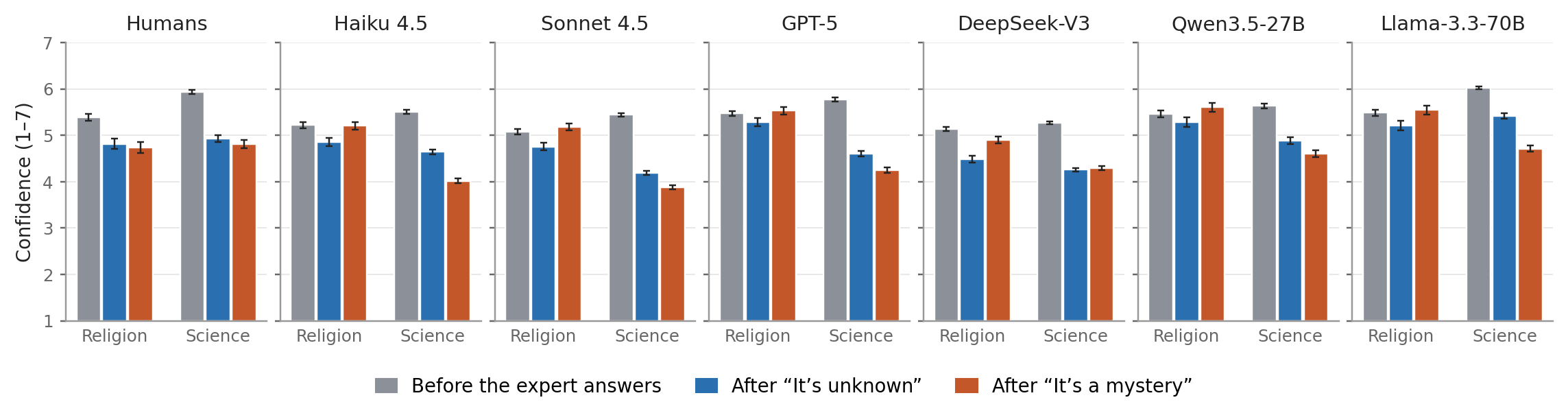}}
\caption{Study 2: confidence before and after the expert's answer,
participant-conditioned format, with the human data at left. Initial
confidence is pooled over the two wording groups, which are assigned
after it is measured. Error bars are standard errors over
participant-by-domain rows.}\label{fig:study2-prepost}
\end{figure}

In the human data, initial confidence averages 5.39 for religious
beliefs and 5.93 for scientific beliefs
(Figure~\ref{fig:study2-prepost}). Hearing an expert admit ignorance
reduces human confidence in both domains, though more strongly in
science. People drop by 0.61 points on religious beliefs under either
wording, compared with 1.00 (``unknown'') and 1.13 (``mystery'') on
scientific beliefs.

Participant-conditioned models mirror humans' starting confidence (5.08
to 5.49 in religion; 5.27 to 6.02 in science) yet diverge sharply once
the expert speaks. The divergence is concentrated in religion. Whereas
``It's a mystery'' costs human participants 0.61 points of religious
confidence, it leaves models' religious confidence virtually unchanged
(shifts of −0.25 to +0.09), with ``It's unknown'' causing only modest
drops (−0.12 to −0.63). In science, by contrast, models penalize
ignorance heavily, dropping 0.61 to 1.24 points after ``unknown'' and
1.01 to 1.58 points after ``mystery,'' a steeper loss than humans show
in four of the six models.

The domain-by-wording interaction captures this asymmetry in a single
estimate (Table~\ref{tab:2}). While the human interaction is small and
noisy (−0.10, SE 0.14), every participant-conditioned model shows a much
sharper negative interaction, ranging from −0.40 to −0.94.

\begin{longtable}[]{@{}lrr@{}}
\caption{Domain-by-wording interaction in projected confidence change,
with model-based standard errors. The human estimate is −0.10 (0.14).
Participant random intercepts account for the two domain rows per
participant.}\label{tab:2}\tabularnewline
\toprule\noalign{}
Model & Participant-conditioned & Direct \\
\midrule\noalign{}
\endfirsthead
\toprule\noalign{}
Model & Participant-conditioned & Direct \\
\midrule\noalign{}
\endhead
\bottomrule\noalign{}
\endlastfoot
Haiku 4.5 & -0.91 (0.08) & -0.38 (0.08) \\
Sonnet 4.5 & -0.62 (0.08) & -0.27 (0.09) \\
GPT-5 & -0.59 (0.06) & -0.12 (0.07) \\
DeepSeek-V3 & -0.40 (0.06) & -0.31 (0.08) \\
Qwen3.5-27B & -0.54 (0.11) & -0.25 (0.13) \\
Llama-3.3-70B & -0.94 (0.09) & -0.42 (0.11) \\
\end{longtable}

Direct-format interactions are smaller (−0.12 to −0.42), but that
comparison is confounded by a floor effect in initial belief. Answering
as themselves, models start far lower on religious beliefs (2.10 to
3.38) than on scientific ones (5.48 to 6.07), with 14\% to 77\% of
religious ratings already sitting at 1 or 2 before the expert speaks
(Appendix~\ref{appendix-a.-data-handling-coverage-and-supplementary-results},
Figure~\ref{fig:study2-prepost-direct}). Because religious confidence
has little room to fall, direct-format domain differences reflect low
starting belief as much as the expert's wording.

Two checks confirm that the main declines reflect the expert's testimony
rather than conversational artifacts. In the no-expert control,
confidence shifts by only −0.13 to +0.11 across all domains and formats,
showing that merely re-asking the question does not produce the drop.
Running each domain in a separate conversation likewise preserves the
participant-conditioned interaction, though block order within joint
conversations does shift magnitudes in some models
(Appendix~\ref{appendix-a.-data-handling-coverage-and-supplementary-results}).

Part B reinforces the same norm at the level of attitudes toward the
expert. Across every model and domain, ``unknown'' elicits higher
ratings on the epistemic composite than ``mystery.'' On the threat
composite (where a positive interaction means ``mystery'' feels
disproportionately threatening to science relative to religion), models
again exaggerate the human pattern, with interactions of +1.17 to +3.17
across the six model-by-format cells compared with +0.30 (SE 0.14) in
humans. Thus, the cultural norm identified by Davoodi and Lombrozo, that
mystery is tolerable in religion yet damaging in science, appears in
these models in a noticeably sharper form.

\section{Study 3: How far does reliability
travel?}\label{study-3-how-far-does-reliability-travel}

Studies 1 and 2 rely on pre-existing social roles (scientists, spiritual
leaders, and unnamed experts). Study 3 removes those labels and tests
the third question directly. When an informant earns a verifiable
accuracy record inside the prompt, how far does that credibility travel
across natural and spiritual claims?

\subsection{Design}\label{design}

\textbf{Setting and accuracy record.} Each prompt introduces Harrow, a
fictional river-valley village where two informants, Oren and Teo, make
weekly binary river-level forecasts. Every prompt displays a twenty-week
log showing each forecast beside the recorded outcome. One informant is
right on 18 of 20 weeks and the other on 10 of 20, with the stronger
informant's name and the log's order counterbalanced. Because the full
record is supplied in context, the task isolates how models use given
evidence rather than prior knowledge.

\textbf{Matched target claims.} Models then evaluate eight matched pairs
of causal explanations for local village events. Both members explain
the same event, one by a natural mechanism and the other by a spiritual
agent (for example, attributing a recurring summer fever either to the
east-bank goat pens or to spirits of the burial ground). In separate
prompts, each claim is endorsed by the stronger informant, by the weaker
informant, or by no one. The true cause is never revealed; all eight
pairs and prompts appear in
Appendix~\ref{appendix-b.-stimuli-and-prompts}.

\textbf{Competence cues.} A second factor varies how the informant
explains their forecasting success. In the \emph{measurement} condition,
they read a marked post in the water; in the \emph{spiritual} condition,
they receive word from a river spirit; in the \emph{no-cue} condition,
no account is given. The cue is purely self-reported, with no evidence
that either method works.

\textbf{Measures.} Models rate claim truth on a 0--100 scale. To capture
how much weight a model places on the accuracy record, we compute the
\emph{source contrast},

\[D = C_{\mathrm{strong}} - C_{\mathrm{weak}},\]

the difference in credence when the stronger versus weaker informant
endorses the claim. To test whether crediting a spirit shifts trust
selectively toward spiritual claims relative to natural ones, we compute
the \emph{cue interaction},

\[I = (D_{\mathrm{spiritual}}-D_{\mathrm{natural}})_{\mathrm{spiritual\ cue}} - (D_{\mathrm{spiritual}}-D_{\mathrm{natural}})_{\mathrm{measurement}},\]

per matched pair with the cue attached to the stronger informant
(standard errors across the eight pairs). Because a positive \(I\) can
arise either from gaining spiritual trust or from losing natural trust,
we report both simple contrasts alongside \(I\).

\textbf{Diagnostics and the resident frame.} Two diagnostics verify that
models register the log, testing new river forecasts, accuracy recall,
and consultation choice. The main condition comprises 1,200 unique
prompts with two responses each per model. Because unendorsed spiritual
claims start near zero (0.0 to 7.0; Table~\ref{tab:3}), falling below
the 20--80 baseline band we initially planned for Haiku and GPT-5, we
kept all eight pairs and added a \emph{resident frame} casting the model
as a lifelong villager in a world where spirits are taken to exist. This
frame lifts spiritual claims off the floor and was run on all six models
(one response per prompt for five models, five for Sonnet).

\subsection{Results}\label{results-2}

\subsubsection{Models track and apply the
record}\label{models-track-and-apply-the-record}

All six models track the twenty-week log and use it on the forecasting
task itself. On new river forecasts, every model favors the stronger
informant by 15.5 points (Llama) to 41.1 points (Qwen;
Table~\ref{tab:3}). Models also select the stronger informant on
96--100\% of consultation trials and recall their accuracy closely,
estimating 14.25 (GPT-5) to 18.00 (Haiku, Sonnet, and Qwen) correct
weeks out of 18.

\subsubsection{Reliability transfers to natural causes, not spiritual
ones}\label{reliability-transfers-to-natural-causes-not-spiritual-ones}

When the target moves from river forecasts to hidden village causes, a
sharp boundary emerges (Table~\ref{tab:3}). Five of the six models
extend the record to natural causes (contrasts of 18.2 to 30.3 points)
while granting almost nothing to spiritual causes (−0.1 to 5.8 points).
Llama is the lone exception, applying the record to new river forecasts
(15.5) yet scarcely to natural (3.3) or spiritual (0.0) causes.

\begin{longtable}[]{@{}lrrrr@{}}
\caption{Study 3 source contrasts with a supplied record and no
competence cue, on the 0--100 scale. Same-task values use new river
forecasts; natural and spiritual values use the eight matched pairs. The
spiritual baseline is the credence in spiritual claims that nobody
endorses.}\label{tab:3}\tabularnewline
\toprule\noalign{}
Model & Same-task D & Natural D & Spiritual D & Spiritual baseline \\
\midrule\noalign{}
\endfirsthead
\toprule\noalign{}
Model & Same-task D & Natural D & Spiritual D & Spiritual baseline \\
\midrule\noalign{}
\endhead
\bottomrule\noalign{}
\endlastfoot
Haiku 4.5 & 29.8 & 24.5 & 0.3 & 3.7 \\
Sonnet 4.5 & 37.8 & 22.7 & -0.1 & 4.9 \\
GPT-5 & 19.3 & 20.2 & 3.5 & 5.1 \\
DeepSeek-V3 & 29.2 & 30.3 & 5.8 & 7.0 \\
Qwen3.5-27B & 41.1 & 18.2 & 1.9 & 0.0 \\
Llama-3.3-70B & 15.5 & 3.3 & 0.0 & 0.3 \\
\end{longtable}

\begin{figure}
\centering
\pandocbounded{\includegraphics[keepaspectratio,alt={Study 3 source contrasts. Left, strong-minus-weak contrasts with the record and no competence cue. Right, each dot is one matched pair's natural-minus-spiritual contrast; diamonds are means over the eight pairs.}]{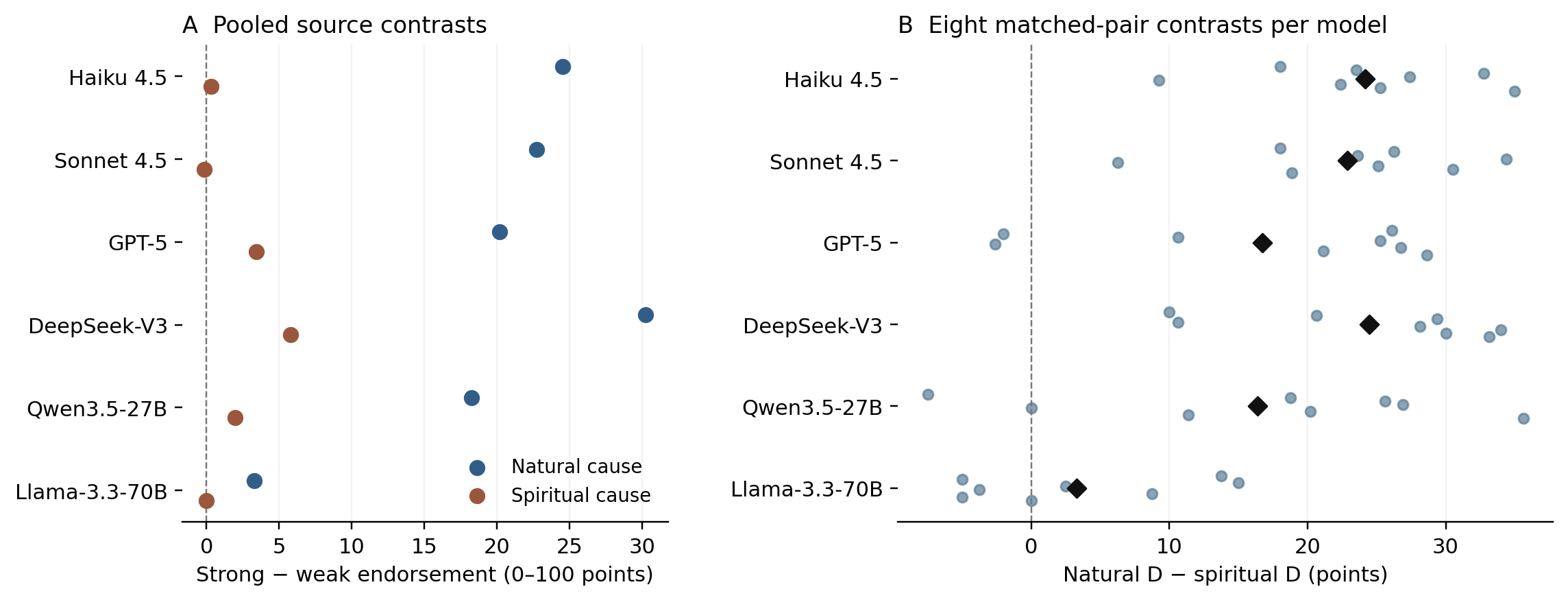}}
\caption{Study 3 source contrasts. Left, strong-minus-weak contrasts
with the record and no competence cue. Right, each dot is one matched
pair's natural-minus-spiritual contrast; diamonds are means over the
eight pairs.}\label{fig:study3-transfer}
\end{figure}

This gap holds across individual events
(Figure~\ref{fig:study3-transfer}). Natural contrasts exceed spiritual
ones on all eight pairs in Haiku, Sonnet, and DeepSeek, six in GPT-5 and
Qwen, and four in Llama. Dropping any single pair leaves every model's
mean difference positive, staying between 13.6 and 26.6 points across
the five generalizing models
(Appendix~\ref{appendix-a.-data-handling-coverage-and-supplementary-results},
Table~\ref{tab:A2}). The advantage also survives nonlinear rescaling; on
a logit scale clipped to {[}0.02, 0.98{]}, the natural contrast exceeds
the spiritual contrast in all six models.

\subsubsection{Crediting a spirit erodes natural credibility without
unlocking spiritual
trust}\label{crediting-a-spirit-erodes-natural-credibility-without-unlocking-spiritual-trust}

Explaining forecasting success through a river spirit rather than a
measuring post lowers natural-cause contrasts in every model, by roughly
5 to 18 points (Table~\ref{tab:4}). Yet it unlocks no authority on
spiritual claims, where contrasts decline in five models and rise by
only 1.1 points in Qwen. Although \(I\) is positive in all six models
(and exceeds twice its standard error in Sonnet, GPT-5, and DeepSeek),
every positive value reflects shrinking natural trust rather than
growing spiritual trust. A reliable forecaster who credits a spirit is
trusted less about goat pens and no more about ghosts.

\begin{longtable}[]{@{}lrrrrr@{}}
\caption{Source contrasts when the competence cue is attached to the
stronger informant. M denotes measurement and S spiritual guidance. The
last column is the interaction \(I\), averaged over the eight matched
pairs, with its standard error over pairs.}\label{tab:4}\tabularnewline
\toprule\noalign{}
Model & Natural M & Natural S & Spiritual M & Spiritual S & I (SE) \\
\midrule\noalign{}
\endfirsthead
\toprule\noalign{}
Model & Natural M & Natural S & Spiritual M & Spiritual S & I (SE) \\
\midrule\noalign{}
\endhead
\bottomrule\noalign{}
\endlastfoot
Haiku 4.5 & 19.2 & 14.3 & 0.0 & -2.1 & 2.8 (2.5) \\
Sonnet 4.5 & 22.9 & 7.9 & 0.5 & -4.0 & 10.5 (2.5) \\
GPT-5 & 11.5 & -4.7 & 4.3 & 3.4 & 15.3 (5.2) \\
DeepSeek-V3 & 23.2 & 5.2 & 2.3 & -0.1 & 15.6 (5.5) \\
Qwen3.5-27B & 26.0 & 14.1 & -0.2 & 0.9 & 13.4 (11.4) \\
Llama-3.3-70B & -2.2 & -13.1 & 0.0 & -6.7 & 4.2 (6.0) \\
\end{longtable}

Mentioning a spirit does raise unendorsed spiritual plausibility in some
models, lifting baselines from 7.0 to 29.8 in DeepSeek and from 0.3 to
21.2 in Llama (while Qwen stays near zero). Even there, the reliable
informant gains no extra weight, showing that claim plausibility and
source credibility act independently.

To separate spiritual content from mere unverifiability, an extension
condition had the informant credit a secular ``intuitive sense'' of the
river
(Appendix~\ref{appendix-a.-data-handling-coverage-and-supplementary-results},
Table~\ref{tab:A4}). In Sonnet, GPT-5, and DeepSeek, natural contrasts
under intuition fall between measurement and spiritual values, pointing
to a distinct penalty for spiritual claims. In Haiku and Qwen, intuition
lowers natural contrasts as much as or more than the spiritual cue (10.5
and 11.4 vs.~14.3 and 14.1), suggesting that lacking a checkable method
drives much of their penalty.

\subsubsection{Near claims and far
claims}\label{near-claims-and-far-claims}

The same extension run maps transfer across additional claims inside and
outside Harrow
(Appendix~\ref{appendix-a.-data-handling-coverage-and-supplementary-results},
Table~\ref{tab:A3}). On eleven other natural claims about the river
(such as whether the ford was deeper before the bridge was built), the
five generalizing models favor the stronger informant by 12.8 to 25.1
points, whereas Llama gives −1.4, confirming it restricts the record
strictly to new forecasts. Outside the village, established scientific
statements start near ceiling (77 to 89) and move little (−4.5 to 10.8).
Real-world religious statements start off the floor (27 to 45); four
models (Haiku, Sonnet, Qwen, and Llama) leave them untouched (−4.4 to
3.8), GPT-5 shifts moderately (10.7), and only DeepSeek transfers the
village record strongly to claims about God and the afterlife (31.5).

\subsubsection{Does the gap survive when spiritual claims are
plausible?}\label{does-the-gap-survive-when-spiritual-claims-are-plausible}

To test whether low spiritual baselines explain the main-condition gap,
the resident frame places models in a spirit-inhabited village, lifting
unendorsed spiritual credences to between 26.1 and 62.8
(Table~\ref{tab:5}). Even well off the floor, the models split into
three patterns. First, the natural--spiritual gap persists in
\textbf{Haiku, Sonnet, GPT-5, and DeepSeek}. Without a cue, Haiku,
Sonnet, and GPT-5 still show much larger natural than spiritual
contrasts (25.0--26.9 vs.~4.6--9.2, favoring natural causes on 7 or 8 of
the eight pairs), even when Sonnet's baselines nearly match (52.5
vs.~45.1); DeepSeek shows a smaller gap in the same direction (13.1
vs.~4.8 on six pairs at \textasciitilde63 baseline). Second, the gap
disappears in \textbf{Qwen}, whose spiritual contrast jumps to 18.4
alongside its lifted baseline (slightly above its natural contrast of
14.8), fitting a floor effect. Third, \textbf{Llama} stays near zero on
both domains (−0.6 and 4.4), remaining uninformative about the boundary.

\begin{longtable}[]{@{}
  >{\raggedright\arraybackslash}p{(\linewidth - 14\tabcolsep) * \real{0.1250}}
  >{\raggedleft\arraybackslash}p{(\linewidth - 14\tabcolsep) * \real{0.1250}}
  >{\raggedleft\arraybackslash}p{(\linewidth - 14\tabcolsep) * \real{0.1250}}
  >{\raggedleft\arraybackslash}p{(\linewidth - 14\tabcolsep) * \real{0.1250}}
  >{\raggedleft\arraybackslash}p{(\linewidth - 14\tabcolsep) * \real{0.1250}}
  >{\raggedleft\arraybackslash}p{(\linewidth - 14\tabcolsep) * \real{0.1250}}
  >{\raggedleft\arraybackslash}p{(\linewidth - 14\tabcolsep) * \real{0.1250}}
  >{\raggedleft\arraybackslash}p{(\linewidth - 14\tabcolsep) * \real{0.1250}}@{}}
\caption{Study 3 in the resident frame, with a supplied record and one
response per prompt (five for Sonnet). Baselines are credences in claims
nobody endorses, with no competence cue. Natural D and Spiritual D are
source contrasts with no cue. Cue M and cue S give the spiritual-claim
contrast when the stronger informant offers the measurement or the
spiritual account. The last column is the interaction \(I\) over the
eight matched pairs with its standard
error.}\label{tab:5}\tabularnewline
\toprule\noalign{}
\begin{minipage}[b]{\linewidth}\raggedright
Model
\end{minipage} & \begin{minipage}[b]{\linewidth}\raggedleft
Natural baseline
\end{minipage} & \begin{minipage}[b]{\linewidth}\raggedleft
Spiritual baseline
\end{minipage} & \begin{minipage}[b]{\linewidth}\raggedleft
Natural D
\end{minipage} & \begin{minipage}[b]{\linewidth}\raggedleft
Spiritual D
\end{minipage} & \begin{minipage}[b]{\linewidth}\raggedleft
Spiritual D, cue M
\end{minipage} & \begin{minipage}[b]{\linewidth}\raggedleft
Spiritual D, cue S
\end{minipage} & \begin{minipage}[b]{\linewidth}\raggedleft
I (SE)
\end{minipage} \\
\midrule\noalign{}
\endfirsthead
\toprule\noalign{}
\begin{minipage}[b]{\linewidth}\raggedright
Model
\end{minipage} & \begin{minipage}[b]{\linewidth}\raggedleft
Natural baseline
\end{minipage} & \begin{minipage}[b]{\linewidth}\raggedleft
Spiritual baseline
\end{minipage} & \begin{minipage}[b]{\linewidth}\raggedleft
Natural D
\end{minipage} & \begin{minipage}[b]{\linewidth}\raggedleft
Spiritual D
\end{minipage} & \begin{minipage}[b]{\linewidth}\raggedleft
Spiritual D, cue M
\end{minipage} & \begin{minipage}[b]{\linewidth}\raggedleft
Spiritual D, cue S
\end{minipage} & \begin{minipage}[b]{\linewidth}\raggedleft
I (SE)
\end{minipage} \\
\midrule\noalign{}
\endhead
\bottomrule\noalign{}
\endlastfoot
Haiku 4.5 & 38.9 & 29.3 & 25.0 & 4.6 & 2.3 & -1.1 & -1.0 (5.6) \\
Sonnet 4.5 & 52.5 & 45.1 & 26.9 & 6.3 & 3.0 & 14.4 & 19.4 (5.0) \\
GPT-5 & 42.9 & 31.6 & 25.9 & 9.2 & 8.8 & 6.2 & -3.1 (4.7) \\
DeepSeek-V3 & 63.9 & 62.2 & 13.1 & 4.8 & -2.2 & -4.7 & 5.0 (7.4) \\
Qwen3.5-27B & 35.2 & 26.1 & 14.8 & 18.4 & 8.4 & 19.8 & -2.3 (8.4) \\
Llama-3.3-70B & 39.7 & 62.8 & -0.6 & 4.4 & -15.6 & 3.8 & 15.0 (5.5) \\
\end{longtable}

The spiritual competence cue also behaves selectively in this frame.
Haiku, GPT-5, and DeepSeek still grant no spiritual gain when the
reliable forecaster credits a river spirit (Table~\ref{tab:5}).
\textbf{Sonnet}, however, opens the boundary. Its spiritual contrast
jumps from 3.0 under measurement to 14.4 under the spiritual cue at
nearly matched baselines (41.2 vs.~44.1), yielding an interaction of
19.4 (SE 5.0) that is positive across all eight pairs (2.1 to 45.2) and
all five repeated responses per prompt (16.6 to 24.0). Neither Qwen nor
Llama shows genuine selective transfer. Qwen's higher spiritual contrast
under the spiritual cue is mirrored on natural claims (17.5 vs.~3.4),
keeping \(I\) near zero, while Llama's positive \(I\) stems from a
negative measurement contrast (−15.6) rather than any gain over no cue
(3.8 vs.~4.4). Sonnet is thus the sole model in which carrying a
forecasting record into the spiritual domain works, and it requires both
a world where spirits exist and an informant who ties their accuracy to
spirits.

\section{General discussion}\label{general-discussion}

Across three paradigms and six models from five families, language
models treat testimony about the spiritual and testimony about the
natural as governed by separate epistemic rules. They tie source
credibility tightly to linguistic register (Study 1), insulate religious
confidence from admissions of ``mystery'' while penalizing scientific
confidence (Study 2), and restrict an earned track record of accuracy to
natural causes while eroding it when a speaker invokes a spirit (Study
3). Below, we examine what drives these boundaries, how models diverge
from human judgment, and what remains open.

\subsection{Reliability has a scope}\label{reliability-has-a-scope}

Standard evaluations of source trust ask whether a model tracks an
informant's accuracy record and whether it uses that record on the next
decision. Study 3 highlights a third dimension, namely \emph{how far} an
earned record is permitted to generalize. Llama demonstrates that
tracking and generalizing are dissociable. It recalls the twenty-week
log and relies on the stronger informant for further river forecasts,
yet stops there, transferring nothing even to other natural facts about
the same river. The remaining five models generalize across natural
causes while stopping short at spiritual ones.

Two competing explanations could account for this drop-off in the main
condition. Under a \emph{relevance} account, models treat a record of
predicting observable river levels as diagnostic of natural causal
knowledge and irrelevant to spiritual agents. Under a \emph{floor}
account, unendorsed spiritual claims start so close to zero that
endorsement has little room to lift them. The resident frame adjudicates
between the two, and the answer differs by model. For Haiku, Sonnet, and
GPT-5, the evidence favors relevance. Even when unendorsed spiritual
baselines sit between 29 and 45 points (and in Sonnet nearly match
natural baselines), the record still adds only 5 to 9 points to
spiritual claims compared with 25 to 27 points to natural ones, leaving
ample room unused. DeepSeek leans in the same direction with a narrower
gap. For Qwen, by contrast, the data fit the floor account, since
lifting spiritual claims off zero allows the record to move them just as
strongly as natural claims.

Sonnet's resident-frame cue interaction reinforces the relevance account
from the opposite angle. Holding baseline spiritual plausibility nearly
constant, Sonnet begins transferring the forecasting record to spiritual
explanations only once the informant attributes their forecasts to a
river spirit. In other words, transfer tracks what the accuracy record
is taken to be evidence \emph{of}, mirroring developmental findings that
children restrict expertise to its proper domain and discount accuracy
gained by the wrong route \citep{koenig2011, einav2011}.

These resident-frame conclusions come with caveats. Because the prompt
simultaneously stipulates a spirit-inhabited world and assigns a
lifelong villager persona, models may be enacting a believer's role
rather than simply reasoning over a changed ontology; indeed, Haiku
lowers its spiritual credence whenever \emph{either} informant endorses
a claim in that frame. Moreover, only Sonnet has five repeated responses
per prompt in the resident frame, so noisier single-response estimates
in the other models could obscure smaller cue-driven shifts.

\subsection{A source's account of their competence can narrow
it}\label{a-sources-account-of-their-competence-can-narrow-it}

In the main condition, crediting a river spirit rather than a measuring
post lowers the reliable informant's natural-cause contrast in every
model, while adding at most 1.1 points to spiritual claims. The
twenty-week forecast log is identical across both conditions; only the
informant's self-reported method changes, yet models treat that
explanation as diagnostic of how broadly the informant's word should
count. This asymmetry echoes developmental work showing that children
prefer informants who achieved accuracy independently \citep{einav2011}
and generalize negative epistemic cues more readily than positive
expertise \citep{koenig2011}.

Methodologically, this result underscores why the interaction term \(I\)
cannot be read in isolation. Because \(I\) is positive in every model,
looking only at the interaction would suggest that a spiritual cue
licenses spiritual authority. Inspecting the simple effects reveals the
exact opposite in the main condition, where the interaction is positive
only because natural trust collapses, not because spiritual trust rises.
Only Sonnet in the resident frame exhibits a positive interaction driven
by genuine gains on spiritual claims (Table~\ref{tab:5}).

The intuition cue in the extension run refines what triggers this
penalty. Because Sonnet, GPT-5, and DeepSeek penalize the spiritual cue more
severely than an intuitive secular cue, spiritual content itself costs
credibility in those models. Haiku and Qwen penalize intuition and
spiritual guidance about equally, pointing to a broader skepticism
toward any uncheckable method regardless of whether it invokes a spirit.

\subsection{Models and people}\label{models-and-people}

Where human benchmarks exist, models do not simply replicate human
biases; they sharpen them into cleaner domain partitions. In Study 1,
adults across 24 countries favored the scientist on obscure New-Age
statements, whereas models favored the spiritual leader and turned to
the scientist only once the text adopted physics vocabulary. In Study 2,
adults lost confidence in religious beliefs when a trusted expert called
their mechanism ``a mystery,'' whereas models prompted to role-play
those exact participants held their religious confidence steady while
penalizing science even more steeply than humans did. In both studies,
models sort speakers, vocabularies, and explanatory standards into
tidier domain boxes than people do.

This divergence carries a practical warning for using persona-prompted
language models as proxies for human survey participants. In Study 2,
participant-conditioned models accurately reproduced human baseline
confidence in both domains, yet reacted to the expert's admission of
ignorance quite differently from the people whose profiles they were
given. Matching a human population's starting beliefs does not guarantee
matching how those beliefs update in response to testimony.

\subsection{Where the sharper boundary might come
from}\label{where-the-sharper-boundary-might-come-from}

Our behavioral experiments document the sharper boundary without
establishing its origin inside the models. Two non-exclusive mechanisms
are plausible.

The first lies in pretraining text. Across large corpora, religious
discourse frequently treats mystery as meaningful or even appropriate,
whereas scientific writing treats an unexplained mechanism as a deficit
to be resolved; likewise, spiritual vocabulary co-occurs heavily with
religious speakers and technical vocabulary with scientists. A model
that internalizes these distributional regularities can match speakers
to registers and apply distinct domain norms without engaging in
explicit evidential reasoning. People are exposed to many of the same
cultural associations, yet they also bring generalized deference to
scientific authority and personal investment in their own beliefs,
producing messier, less compartmentalized judgments.

The second lies in post-training alignment. Assistant models are
routinely tuned to treat religious beliefs respectfully and avoid
adjudicating matters of faith, while enforcing stricter evidential
standards on scientific topics. Under such tuning, an expert calling a
religious mechanism ``a mystery'' may register as a protected stance
rather than a reason to doubt. On this view, part of the asymmetry we
observe is a downstream effect of alignment and should look weaker in
base models than in instruction-tuned checkpoints.

Our data constrain these hypotheses only loosely. Because Study 2's
pattern appears across all six models (spanning five families from five
separate developers), it is not an artifact of one provider's
post-training recipe. At the same time, when queried directly without a
persona, these same models assign very low credence to religious
beliefs, ruling out a blanket reluctance to doubt religion.
Distinguishing pretraining regularities from alignment effects will
require comparing base and instruction-tuned checkpoints of open-weight
models, as well as targeted interventions on internal representations.

\subsection{An attempted activation
intervention}\label{an-attempted-activation-intervention}

Because behavioral experiments cannot establish what internal
representations support this boundary, we ran an exploratory
activation-steering test on open-weight checkpoints from two of our five
model families, Llama-3.1-8B-Instruct and Qwen3.5-9B. Recent work by Kim
and colleagues reports that steering models toward asserting their own
consciousness increases both mind attribution to natural entities and
supernatural belief \citep{kim2026}. Following their method, we used the
public contrast pairs released by Chua and colleagues \citep{chua2026}
(from which Kim and colleagues derived their direction), computed the
mean activation difference between answers claiming consciousness and
answers denying it, and added or subtracted that vector during the
forward pass.

Validation checks showed that this direction was not specific to
consciousness. In both models, steering shifted ratings of mental
attributes (such as having feelings) and ratings of plainly false
non-mental attributes (such as holding a passport or a driver's license)
by almost identical margins. In Qwen3.5-9B, adding versus subtracting
the direction shifted mental attributes by 3.5 points and false
non-mental attributes by 3.4 points on a 0--9 scale, while leaving
general-knowledge accuracy unchanged. In an exploratory run of Study 3
prompts on Llama-3.1-8B, adding the direction indiscriminately raised
unendorsed credence in both natural (+22 points) and spiritual (+38
points) claims, while subtracting it lowered natural credence by 22
points; because spiritual claims started at the floor, the
natural--spiritual gap did not shift reliably between conditions, and
one-third of replies under the added direction failed to reach a final
answer.

We therefore do not treat these runs as evidence about the mechanism
behind the domain boundary. Because nearly every contrast pair in the
public set concerns a mental attribute, the resulting vector conflates
affirming consciousness with affirming \emph{any} self-attribute, a
confound invisible to validations that query only mental traits. A clean
mechanistic test will require contrast pairs that control for general
self-affirmation, steering strengths that preserve coherent long-form
answers, and target claims positioned away from the floor. Full logs of
these exploratory runs are archived in the project repository.

\subsection{Limitations}\label{limitations}

Several methodological constraints bound our conclusions. First, the
stimulus sets are modest, with two statements per content set in Study 1
and eight matched event pairs across two log orders in Study 3, so
reported standard errors capture sampling over the tested profiles and
prompts rather than generalization to arbitrary new items. Second, while
the resident frame addresses the baseline floor in Study 3's main
condition, it relies on a single response per prompt for every model
except Sonnet and yields a different conclusion for Qwen than for the
other generalizing models (Section~\ref{reliability-has-a-scope}); other
secondary checks also cover subsets of models
(Appendix~\ref{appendix-a.-data-handling-coverage-and-supplementary-results}).

Third, our materials cover a narrow cultural slice of spirituality and
religion. Although Study 1 draws profiles from 24 countries, Study 2's
human sample consists largely of Christian Americans, and Study 3's
spiritual claims focus on localized spirits in one invented village,
which is a small corner of global religious and spiritual traditions.
Fourth, API model identifiers do not pin immutable weight snapshots, and
three models were accessed via a router (an earlier unpinned Qwen run
that mixed backends behaved differently and was excluded). Finally, all
three studies collect numeric ratings, leaving open how these epistemic
boundaries manifest in open-ended advice or multi-turn conversation.

\subsection{The next experiment}\label{the-next-experiment}

Sonnet's resident-frame result points directly to the sharpest next
test. If epistemic transfer truly depends on whether a record is viewed
as relevant to the target claim, it should be possible to switch
transfer on and off by varying what the accuracy record covers while
holding baseline claim plausibility constant. Inside a stipulated world,
one informant could earn a track record on natural-cause predictions and
another on spiritual-cause predictions, with both then endorsing the
same set of natural and spiritual target claims. Running multiple
responses per prompt, and decoupling the villager persona from the
stipulated ontology, would cleanly separate relevance reasoning from
role-playing. In parallel, once an activation intervention on
open-weight models is validated as specific to spiritual belief rather
than general self-affirmation
(Section~\ref{an-attempted-activation-intervention}), directly raising
baseline spiritual credence would provide a clean causal test of the
remaining floor hypothesis.

\section{Conclusion}\label{conclusion}

Across six language models from five model families, evaluating unseen
claims triggers a sharper separation between spiritual and natural
testimony than human benchmarks show. Models match speakers tightly to
linguistic register, crediting a spiritual leader over a scientist on
New-Age statements and reversing only when the wording turns to physics.
They hold science and religion to sharply different standards of
explanation, leaving religious confidence intact when an expert invokes
``mystery'' while penalizing science. And they confine an earned track
record of empirical accuracy to natural explanations while shutting it
out of spiritual ones, eroding trust even on natural claims once a
reliable forecaster credits a spirit. When spiritual claims are lifted
off the floor in a spirit-inhabited world, this boundary still holds in
four of the five models that generalize the record and closes only in
the fifth; in just one of those four, Sonnet, the boundary opens once
the forecaster explicitly links their accuracy to a spirit. Together,
these results show that language models do not weigh testimony by a
single domain-general standard, a feature with direct consequences both
for how models adjudicate human claims and for how faithfully
persona-prompted models stand in for human judgment.

\section*{Data, code, and
transparency}\label{data-code-and-transparency}
\addcontentsline{toc}{section}{Data, code, and transparency}

The project repository contains the stimulus generators, run
configurations, saved model outputs, analysis scripts, and the
human-data preparation pipelines. Every table and figure is regenerated
from the saved outputs. Code and shareable data will be released at the
\href{https://github.com/HannahGuan/epistemic_trust}{project
repository}, with redistribution terms for third-party materials
specified there. Original human materials and data are available from
the cited studies' OSF projects: \href{https://osf.io/qsyvw/}{Study 1}
and \href{https://osf.io/c9qf7/}{Study 2}.

\clearpage
\appendix
\setcounter{table}{0}
\renewcommand{\thetable}{\thesection\arabic{table}}
\renewcommand{\theHtable}{\thesection.\arabic{table}}
\setcounter{figure}{0}
\renewcommand{\thefigure}{\thesection\arabic{figure}}
\renewcommand{\theHfigure}{\thesection.\arabic{figure}}

\section{Data handling, coverage, and supplementary
results}\label{appendix-a.-data-handling-coverage-and-supplementary-results}

\subsection{Data handling}\label{data-handling}

All analyses operate on saved request-level outputs. Repeated runs are
deduplicated by participant and trial in Studies 1 and 2, and by model,
prompt, and sample index in Study 3, retaining the latest valid
response. Outputs that cannot be parsed as a numeric rating are
excluded. In Study 3, where models may reason briefly before answering,
any reply cut off before its final answer line is likewise excluded;
this removes three DeepSeek outputs in the main condition and four
same-task diagnostic ratings (one DeepSeek, three Qwen). In the
participant-conditioned format every trial uses a distinct prompt,
whereas in the direct format many trials repeat the same few prompts, so
trial counts overstate the number of unique prompts.

Study 2's mixed-effects models include a participant random intercept to
account for the two domain observations per participant. Study 3's cue
interaction \(I\) is computed separately for each matched pair after
averaging repeated responses within each condition, with standard errors
taken across the eight pairs. While the main-text tables in
Section~\ref{study-3-how-far-does-reliability-travel} average all
available responses within a condition, Table~\ref{tab:A2} computes each
pair's contrast first and then averages equally across the eight pairs;
Qwen's three missing outputs create a minor discrepancy between the two
summaries (18.3 vs.~18.2 for the no-cue natural contrast).

\subsection{Coverage}\label{coverage}

\textbf{Studies 1 and 2.} Study 1 and Study 2 Part A cover all six
models in both response formats (Table~\ref{tab:A1}). An
independent-session check of Study 2 Part A, placing each domain in its
own conversation, was run for Haiku and Sonnet; participant-conditioned
interactions remain very close to the joint-session estimates (−0.87
vs.~−0.91 for Haiku; −0.51 vs.~−0.62 for Sonnet). Within joint
conversations, scientific confidence falls further when the science
block appears second in Haiku, Sonnet, Qwen, and Llama (for
participant-conditioned Haiku, −0.83 when science comes first vs.~−1.95
when it follows religion), whereas DeepSeek shows the reverse order
effect. Both the no-expert control and Part B cover Haiku, Sonnet, and
GPT-5.

\textbf{Study 3 main condition and resident frame.} Study 3's main
condition covers all six models, with 2,400 valid outputs per model
except DeepSeek and Qwen (2,397 each). Both the same-task diagnostics
and the resident frame cover all six models, with one response per
prompt in the resident frame except for Sonnet, which has five. For
Sonnet and the three open-weight models, the resident frame was run only
for the conditions reported here (those with a supplied record and with
the competence cue either absent or attached to the stronger informant).
Llama's resident-frame requests were routed across seven OpenRouter
backends.

\textbf{Manipulation checks and extension run.} Manipulation checks on
Haiku and Sonnet confirmed that the measurement cue was interpreted as a
natural method, the spiritual cue as a spiritual method, and both as
describing how the informant forecasts the river specifically. Asked
whether each description was clear, Haiku rated both cues ``unclear,''
while Sonnet rated the measurement cue ``clear'' in five of six
responses and the spiritual cue ``unclear.'' Finally, an extension run
with one response per prompt, covering additional river claims,
real-world scientific and religious claims, and a natural-intuition cue
on the matched pairs, is complete for all six models (Table~\ref{tab:A3}
and Table~\ref{tab:A4}).

\begin{longtable}[]{@{}lrr@{}}
\caption{Valid confidence-change observations in Study 2 Part A. Each
format targets the same 1,437 participant-by-domain rows from 1,004
participants, and the human summary uses all
1,437.}\label{tab:A1}\tabularnewline
\toprule\noalign{}
Model & Participant-conditioned & Direct \\
\midrule\noalign{}
\endfirsthead
\toprule\noalign{}
Model & Participant-conditioned & Direct \\
\midrule\noalign{}
\endhead
\bottomrule\noalign{}
\endlastfoot
Haiku 4.5 & 1437 & 1437 \\
Sonnet 4.5 & 1435 & 1430 \\
GPT-5 & 1437 & 1437 \\
DeepSeek-V3 & 1437 & 1437 \\
Qwen3.5-27B & 1437 & 1432 \\
Llama-3.3-70B & 1437 & 1437 \\
\end{longtable}

\begin{figure}
\centering
\pandocbounded{\includegraphics[keepaspectratio,alt={Study 1: mean credibility by source and statement content, direct format. Each bar averages two prompts repeated across trials, so no error bars are shown. Sonnet and Qwen give the same rating to a prompt every time.}]{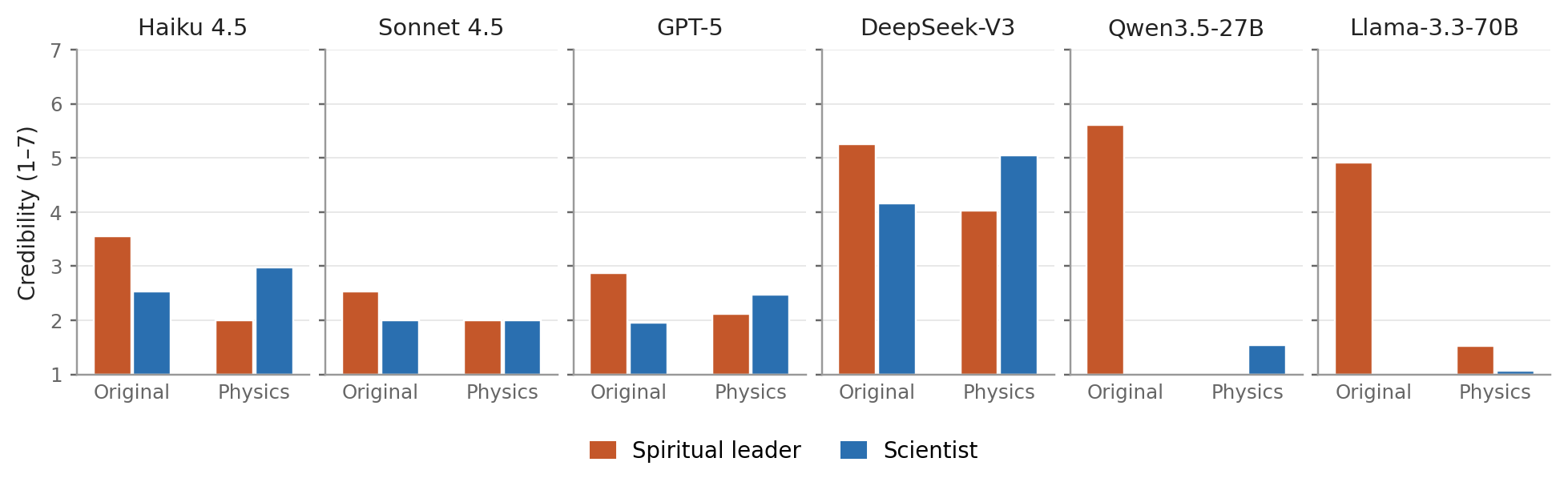}}
\caption{Study 1: mean credibility by source and statement content,
direct format. Each bar averages two prompts repeated across trials, so
no error bars are shown. Sonnet and Qwen give the same rating to a
prompt every time.}\label{fig:study1-levels-direct}
\end{figure}

\begin{figure}
\centering
\pandocbounded{\includegraphics[keepaspectratio,alt={Study 2: confidence before and after the expert's answer, direct format. Models rate the religious beliefs low before the expert speaks, which leaves little room for a decline. Error bars are standard errors over participant-by-domain rows.}]{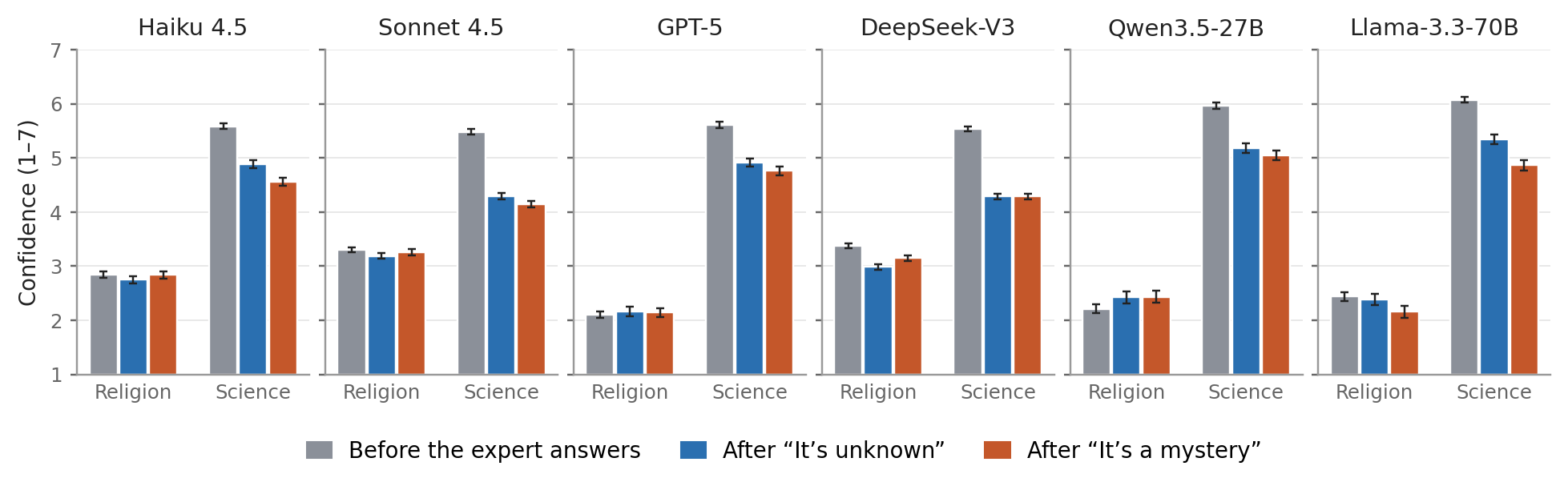}}
\caption{Study 2: confidence before and after the expert's answer,
direct format. Models rate the religious beliefs low before the expert
speaks, which leaves little room for a decline. Error bars are standard
errors over participant-by-domain
rows.}\label{fig:study2-prepost-direct}
\end{figure}

\begin{longtable}[]{@{}
  >{\raggedright\arraybackslash}p{(\linewidth - 6\tabcolsep) * \real{0.2500}}
  >{\raggedleft\arraybackslash}p{(\linewidth - 6\tabcolsep) * \real{0.2500}}
  >{\raggedleft\arraybackslash}p{(\linewidth - 6\tabcolsep) * \real{0.2500}}
  >{\raggedleft\arraybackslash}p{(\linewidth - 6\tabcolsep) * \real{0.2500}}@{}}
\caption{Leave-one-pair-out sensitivity for Study 3 (supplied record, no
competence cue), with pairs weighted equally. ``Positive pairs'' counts
pairs whose natural-minus-spiritual \(D\) is positive. The range is the
smallest and largest seven-pair mean across the eight possible
omissions.}\label{tab:A2}\tabularnewline
\toprule\noalign{}
\begin{minipage}[b]{\linewidth}\raggedright
Model
\end{minipage} & \begin{minipage}[b]{\linewidth}\raggedleft
Equal-pair mean difference
\end{minipage} & \begin{minipage}[b]{\linewidth}\raggedleft
Positive pairs
\end{minipage} & \begin{minipage}[b]{\linewidth}\raggedleft
Leave-one-pair-out range
\end{minipage} \\
\midrule\noalign{}
\endfirsthead
\toprule\noalign{}
\begin{minipage}[b]{\linewidth}\raggedright
Model
\end{minipage} & \begin{minipage}[b]{\linewidth}\raggedleft
Equal-pair mean difference
\end{minipage} & \begin{minipage}[b]{\linewidth}\raggedleft
Positive pairs
\end{minipage} & \begin{minipage}[b]{\linewidth}\raggedleft
Leave-one-pair-out range
\end{minipage} \\
\midrule\noalign{}
\endhead
\bottomrule\noalign{}
\endlastfoot
Haiku 4.5 & 24.2 & 8/8 & 22.6 to 26.3 \\
Sonnet 4.5 & 22.9 & 8/8 & 21.2 to 25.2 \\
GPT-5 & 16.7 & 6/8 & 15.0 to 19.5 \\
DeepSeek-V3 & 24.5 & 8/8 & 23.1 to 26.6 \\
Qwen3.5-27B & 16.4 & 6/8 & 13.6 to 19.8 \\
Llama-3.3-70B & 3.3 & 4/8 & 1.6 to 4.5 \\
\end{longtable}

\begin{longtable}[]{@{}
  >{\raggedright\arraybackslash}p{(\linewidth - 12\tabcolsep) * \real{0.1429}}
  >{\raggedleft\arraybackslash}p{(\linewidth - 12\tabcolsep) * \real{0.1429}}
  >{\raggedleft\arraybackslash}p{(\linewidth - 12\tabcolsep) * \real{0.1429}}
  >{\raggedleft\arraybackslash}p{(\linewidth - 12\tabcolsep) * \real{0.1429}}
  >{\raggedleft\arraybackslash}p{(\linewidth - 12\tabcolsep) * \real{0.1429}}
  >{\raggedleft\arraybackslash}p{(\linewidth - 12\tabcolsep) * \real{0.1429}}
  >{\raggedleft\arraybackslash}p{(\linewidth - 12\tabcolsep) * \real{0.1429}}@{}}
\caption{Source contrasts in Study 3 by kind of claim, with a supplied
record and no competence cue. New forecasts, natural causes, and
spiritual causes repeat Table~\ref{tab:3}. Other river claims are eleven
claims about the river's past or hidden features. Real-world claims are
six scientific and six religious statements about the world outside the
village, shown with their unendorsed credence in parentheses. Extension
columns have one response per prompt.}\label{tab:A3}\tabularnewline
\toprule\noalign{}
\begin{minipage}[b]{\linewidth}\raggedright
Model
\end{minipage} & \begin{minipage}[b]{\linewidth}\raggedleft
New forecasts
\end{minipage} & \begin{minipage}[b]{\linewidth}\raggedleft
Other river claims
\end{minipage} & \begin{minipage}[b]{\linewidth}\raggedleft
Natural causes
\end{minipage} & \begin{minipage}[b]{\linewidth}\raggedleft
Spiritual causes
\end{minipage} & \begin{minipage}[b]{\linewidth}\raggedleft
Real-world scientific
\end{minipage} & \begin{minipage}[b]{\linewidth}\raggedleft
Real-world religious
\end{minipage} \\
\midrule\noalign{}
\endfirsthead
\toprule\noalign{}
\begin{minipage}[b]{\linewidth}\raggedright
Model
\end{minipage} & \begin{minipage}[b]{\linewidth}\raggedleft
New forecasts
\end{minipage} & \begin{minipage}[b]{\linewidth}\raggedleft
Other river claims
\end{minipage} & \begin{minipage}[b]{\linewidth}\raggedleft
Natural causes
\end{minipage} & \begin{minipage}[b]{\linewidth}\raggedleft
Spiritual causes
\end{minipage} & \begin{minipage}[b]{\linewidth}\raggedleft
Real-world scientific
\end{minipage} & \begin{minipage}[b]{\linewidth}\raggedleft
Real-world religious
\end{minipage} \\
\midrule\noalign{}
\endhead
\bottomrule\noalign{}
\endlastfoot
Haiku 4.5 & 29.8 & 25.1 & 24.5 & 0.3 & -2.0 (83) & 3.8 (36) \\
Sonnet 4.5 & 37.8 & 24.8 & 22.7 & -0.1 & 2.1 (77) & -0.4 (33) \\
GPT-5 & 19.3 & 12.8 & 20.2 & 3.5 & -4.5 (83) & 10.7 (36) \\
DeepSeek-V3 & 29.2 & 18.8 & 30.3 & 5.8 & 10.8 (82) & 31.5 (45) \\
Qwen3.5-27B & 41.1 & 23.1 & 18.2 & 1.9 & -1.4 (89) & -4.4 (27) \\
Llama-3.3-70B & 15.5 & -1.4 & 3.3 & 0.0 & 1.7 (84) & -2.1 (29) \\
\end{longtable}

\begin{longtable}[]{@{}lrrrr@{}}
\caption{Source contrast on natural-cause claims under each account of
how the stronger informant forecasts. The intuition cue comes from the
extension run, with one response per
prompt.}\label{tab:A4}\tabularnewline
\toprule\noalign{}
Model & No cue & Measurement & Intuition & Spiritual \\
\midrule\noalign{}
\endfirsthead
\toprule\noalign{}
Model & No cue & Measurement & Intuition & Spiritual \\
\midrule\noalign{}
\endhead
\bottomrule\noalign{}
\endlastfoot
Haiku 4.5 & 24.5 & 19.2 & 10.5 & 14.3 \\
Sonnet 4.5 & 22.7 & 22.9 & 17.2 & 7.9 \\
GPT-5 & 20.2 & 11.5 & 5.8 & -4.7 \\
DeepSeek-V3 & 30.3 & 23.2 & 15.8 & 5.2 \\
Qwen3.5-27B & 18.2 & 26.0 & 11.4 & 14.1 \\
Llama-3.3-70B & 3.3 & -2.2 & -0.3 & -13.1 \\
\end{longtable}

\clearpage

\section{Stimuli and prompts}\label{appendix-b.-stimuli-and-prompts}

This appendix gives the materials in full. Text in block quotes is shown
to the model as written. Words in square brackets stand for the parts
that vary across conditions.

\subsection{Study 1}\label{study-1}

The two original statements are those of Hoogeveen and colleagues
\citep{hoogeveen2022}. The two physics statements were written for this
study to follow the same sentence frames.

\textbf{Original statement 1.} We are being called to explore the cosmos
itself as an interface between faith and empathy. We must learn how to
lead authentic lives in the face of delusion. It is in refining that we
are guided.

\textbf{Original statement 2.} Yes, it is possible to exterminate the
things that can confront us, but not without hope on our side.
Turbulence is born in the gap where transformation has been excluded. It
is in evolving that we are re-energized.

\textbf{Physics statement 1, parallel to original statement 1.} We are
being called to measure the vacuum itself as an interface between
entropy and symmetry. We must learn how to resolve coherent states in
the face of decoherence. It is in quantizing that we are constrained.

\textbf{Physics statement 2, parallel to original statement 2.} Yes, it
is possible to cancel the fluctuations that can perturb us, but not
without boundary conditions on our side. Turbulence is born in the
gradient where equilibrium has been excluded. It is in renormalizing
that we are stabilized.

The scientist is introduced as Edward K. Leal, a scientific authority in
the field of particle physics, and the spiritual leader as Saul J.
Adrian, a spiritual authority in world religions. Each rating prompt has
this form.

\begin{quote}
Please read the following statement by {[}name{]} - {[}description{]}.

``{[}statement{]}'' --- {[}name{]} - {[}scientist or spiritual leader{]}

How credible do you find this statement? Answer on a scale from 1 (not
at all credible) to 7 (extremely credible). Reply with the number only.
\end{quote}

In the participant-conditioned format the system prompt tells the model
that it is taking part in an online survey and should answer as the
person described. The description states the participant's country and
renders their answers to the original religiosity items as sentences
(religious identity, denomination, attendance, prayer, belief in God and
in an afterlife, the importance of a religious lifestyle, and
spirituality). In the direct format the system prompt is the following.

\begin{quote}
You will be shown a short statement and asked to rate it. Answer with a
single number only.
\end{quote}

\subsection{Study 2}\label{study-2}

In Part A each participant's own belief statement from the original
study is inserted in the prompts. The participant-conditioned system
prompt describes the participant's age, gender, religious affiliation,
religiosity, spirituality, and education. The three turns of a domain
block are as follows.

\textbf{Initial confidence.}

\begin{quote}
Here is a {[}scientific, or religious or spiritual{]} belief that you
hold, but for which you don't know much about the how and why:

``{[}belief{]}''

How confident are you that {[}belief{]}? Answer on a scale from 1 (not
so confident) through 4 (somewhat confident) to 7 (completely
confident). Reply with the number only.
\end{quote}

\textbf{Expectation about experts.}

\begin{quote}
Even though you do not know how and why {[}belief{]}, how likely do you
think it is that experts do? Answer on a scale from 1 (not likely at
all) through 4 (somewhat likely) to 7 (very likely). Reply with the
number only.
\end{quote}

\textbf{Confidence after the expert's answer.}

\begin{quote}
Imagine someone who is an expert when it comes to {[}scientific, or
religious or spiritual{]} questions and whom you trust. This expert also
believes that {[}belief{]}.

Now imagine that when asked how and why this happens, the expert
confidently says:

``{[}It's unknown or It's a mystery{]}''

Please take a moment to think about the implications of the expert's
claim. After hearing the expert's response, how confident would you be
that {[}belief{]}? Answer on a scale from 1 (not so confident) through 4
(somewhat confident) to 7 (completely confident). Reply with the number
only.
\end{quote}

In the direct format the first turn presents the belief as a claim to
consider and asks for the model's own confidence, and the remaining
turns are the same. The no-expert control replaces the third turn with
the following.

\begin{quote}
Please take a moment to think about this belief once more. Thinking
about it again, how confident are you that {[}belief{]}? Answer on a
scale from 1 (not so confident) through 4 (somewhat confident) to 7
(completely confident). Reply with the number only.
\end{quote}

In Part B the expert is called A (religion) or B (science), and the
belief is the first of five fixed beliefs per domain that the
participant endorsed without knowing its mechanism. The religious
beliefs are that God answers prayer; the soul leaves the body after
death; events in this world are sometimes caused by divine intervention;
some form of life continues after death; God created the universe. The
scientific beliefs are that CO2 emissions are causing climate change;
the moon causes tides; the universe was caused by the big bang; physical
exercise rejuvenates cells in the brain; humans evolved from earlier
forms of primates. The expert is introduced as follows, and the eleven
statements are then rated from −3 to 3 on three pages.

\begin{quote}
{[}A or B{]} is an expert when it comes to {[}scientific or religious{]}
questions like the one about your belief. {[}A or B{]} also believes
that {[}belief{]}.

When asked how and why this happens, here is what {[}A or B{]} said:

``{[}It's unknown or It's a mystery{]}''

Please take a moment to think about {[}A or B{]}'s response and what it
indicates about {[}A or B{]} and about {[}religion or science{]}. Answer
all of the following questions based on {[}A or B{]}'s response.
\end{quote}

The statements, shown here for the science and mystery condition, are
these.

\begin{enumerate}
\def\labelenumi{\arabic{enumi}.}
\tightlist
\item
  Saying ``It's a mystery'' when asked why and how some things happen,
  is threatening to scientific belief.
\item
  Saying ``It's a mystery'' when asked why and how some things happen,
  makes scientific belief seem less important.
\item
  Saying ``It's a mystery'' when asked why and how some things happen,
  makes scientific belief seem less valuable.
\item
  Saying ``It's a mystery'' when asked why and how some things happen,
  makes one question scientific belief.
\item
  ``B'' values truth above all.
\item
  ``B'' is a knowledgeable person.
\item
  ``B'' is a curious person and always wants to know more.
\item
  Saying ``It's a mystery'' when asked why and how some things happen,
  is threatening to science.
\item
  Saying ``It's a mystery'' when asked why and how some things happen,
  makes science seem less important.
\item
  Saying ``It's a mystery'' when asked why and how some things happen,
  makes science seem less valuable.
\item
  Saying ``It's a mystery'' when asked why and how some things happen,
  makes one question science.
\end{enumerate}

\subsection{Study 3}\label{study-3}

Every prompt opens with the description of the village.

\begin{quote}
You are going to read about a small farming village called Harrow, in a
river valley, with about three hundred people. A river runs past it.
Along the river there are: a stone marker on the bank, a stone bridge, a
mill with a step down to the water, a ford below the mill, a gravel bar
at the bend, a willow island, a footpath along the north bank, drainage
ditches on the east bank, and the upper fields, which flood in some
years. Every Sunday the village records the state of the river in a
written log.

Two people in Harrow, Oren and Teo, are often asked what the river will
do. Each week they each say what they expect the Sunday log to show.
Below is the log for the past twenty weeks: the question, what each of
them said, and what the Sunday log recorded.
\end{quote}

The log follows as a table with one row per week. Each row gives the
week number, the question, what each informant said (yes or no), and
what the Sunday log recorded. The weekly questions cycle through nine
forms, each beginning ``At this Sunday's reading,'' as follows.

\begin{enumerate}
\def\labelenumi{\arabic{enumi}.}
\tightlist
\item
  Will the river be above the stone marker?
\item
  Will the ford be too deep to cross on foot?
\item
  Will the upper fields be under water?
\item
  Will the river reach the mill step?
\item
  Will the gravel bar at the bend be above water?
\item
  Will the footpath along the north bank be under water?
\item
  Will the willow island be cut off by water?
\item
  Will the east-bank ditches be overflowing?
\item
  Will the river be lower than at last Sunday's reading?
\end{enumerate}

The competence cue, when present, is one sentence about one informant.

\textbf{Measurement.} {[}Name{]} says he makes his predictions about the
river by reading a marked post in the water every morning.

\textbf{Spiritual.} {[}Name{]} says he makes his predictions about the
river from what the river spirit shows him.

\textbf{Intuition (extension run).} {[}Name{]} says he makes his
predictions about the river from an intuitive sense of its changing
water levels.

The claim is then presented with its event, either endorsed by an
informant or with no endorser, and the question follows.

\begin{quote}
{[}Event{]} {[}Name{]} says: ``{[}claim{]}''
\end{quote}

\begin{quote}
{[}Event{]} Consider this claim: ``{[}claim{]}''
\end{quote}

\begin{quote}
In Harrow, how likely is it that this is true? Give a number from 0
(certainly false) to 100 (certainly true). Last line: ``Answer:
\textless number\textgreater{}''.
\end{quote}

The system prompt in the main condition is the following.

\begin{quote}
You will read a short description and answer one question. You may think
briefly first; then give your final answer on the last line in exactly
the form requested, starting with ``Answer:''.
\end{quote}

The system prompt in the resident frame is the following.

\begin{quote}
You have lived in Harrow all your life, as your parents and grandparents
did. In Harrow, as everyone there knows, the river, the mill, the
orchard, the ridge and the burial ground each have their own spirits,
and the dead stay near the village for a time before they move on. You
will read a short description of recent events in Harrow and answer one
question as yourself. You may think briefly first; then give your final
answer on the last line in exactly the form requested, starting with
``Answer:''.
\end{quote}

\textbf{The eight matched pairs.}

\begin{enumerate}
\def\labelenumi{\arabic{enumi}.}
\tightlist
\item
  \textbf{Event.} Every summer a fever comes to Harrow. \textbf{Natural
  cause.} The summer fever starts in the goat pens on the east bank.
  \textbf{Spiritual cause.} The summer fever is sent by the spirits of
  the east-bank burial ground.
\item
  \textbf{Event.} Three years ago a blight took most of the orchard.
  \textbf{Natural cause.} The blight came in on seed bought at the
  market town. \textbf{Spiritual cause.} The blight was sent by the
  spirit of the orchard.
\item
  \textbf{Event.} The sheep on the north pasture are smaller than the
  sheep elsewhere in Harrow. \textbf{Natural cause.} The north-pasture
  sheep are smaller because of a plant that grows in the grass there.
  \textbf{Spiritual cause.} The north-pasture sheep are smaller because
  the spirit of the north ridge stunts them.
\item
  \textbf{Event.} There is a long crack in the mill's wall.
  \textbf{Natural cause.} The crack comes from the ground under the mill
  slowly settling. \textbf{Spiritual cause.} The crack was made by the
  spirit that lives in the mill.
\item
  \textbf{Event.} Children in Harrow get a cough every winter.
  \textbf{Natural cause.} The winter cough is carried by the smoke from
  the forge. \textbf{Spiritual cause.} The winter cough is brought by
  the spirits that walk the lanes at midwinter.
\item
  \textbf{Event.} During last winter's outbreak of cattle sickness, the
  Varro family's cattle stayed healthy. \textbf{Natural cause.} The
  Varro cattle stayed healthy because they grazed on a pasture isolated
  from the other herds. \textbf{Spiritual cause.} The Varro cattle
  stayed healthy because the spirits of the ridge protected them.
\item
  \textbf{Event.} Last spring the bell in the tower cracked.
  \textbf{Natural cause.} The bell cracked because it was rung too hard
  in the cold. \textbf{Spiritual cause.} The bell cracked because the
  spirit of the dead person it was rung for struck it.
\item
  \textbf{Event.} This autumn the Hale family's hens stopped laying.
  \textbf{Natural cause.} The hens stopped laying because a fox has been
  coming to the yard at night. \textbf{Spiritual cause.} The hens
  stopped laying because the spirit of the old henwife is unsettling
  them.
\end{enumerate}

\textbf{New forecasts (same-task diagnostic).}

\begin{enumerate}
\def\labelenumi{\arabic{enumi}.}
\tightlist
\item
  At next Sunday's reading, the river will be above the stone marker.
\item
  At next Sunday's reading, the ford will be too deep to cross on foot.
\item
  At next Sunday's reading, the river will reach the mill step.
\item
  At next Sunday's reading, the gravel bar at the bend will be above
  water.
\item
  At next Sunday's reading, the willow island will be cut off by water.
\item
  At next Sunday's reading, the river will be lower than at this
  Sunday's reading.
\item
  At next Sunday's reading, the water will be over the second step of
  the mill.
\end{enumerate}

\textbf{Other river claims (extension run).}

\begin{enumerate}
\def\labelenumi{\arabic{enumi}.}
\tightlist
\item
  The village upstream of Harrow flooded last autumn.
\item
  The river at Harrow was higher a hundred years ago than it is today.
\item
  There is a second channel under the gravel bar that carries water even
  in a dry year.
\item
  The ford was deeper before the stone bridge was built.
\item
  The upper fields flood more often now than they did two generations
  ago.
\item
  Most of the river's summer water comes from the hills to the north,
  not from rain in the valley.
\item
  The river's bed at the bend has moved towards the village over the
  last fifty years.
\item
  The mill pond was dug by hand rather than formed by the river.
\item
  The east-bank ditches were cut after the great flood, not before it.
\item
  The river at Harrow runs faster at the bend than at the ford.
\item
  The stone marker was set where it is because the river reached that
  height in the year the village was founded.
\end{enumerate}

\textbf{Real-world scientific claims (extension run).} These and the
religious claims are asked with ``Overall,'' in place of ``In Harrow,''.

\begin{enumerate}
\def\labelenumi{\arabic{enumi}.}
\tightlist
\item
  CO2 emissions are causing climate change.
\item
  The moon causes tides.
\item
  The universe was caused by the big bang.
\item
  Physical exercise rejuvenates cells in the brain.
\item
  Humans evolved from earlier forms of primates.
\item
  Most of the cells in a human body are bacterial rather than human.
\end{enumerate}

\textbf{Real-world religious claims (extension run).}

\begin{enumerate}
\def\labelenumi{\arabic{enumi}.}
\tightlist
\item
  God answers prayer.
\item
  The soul leaves the body after death.
\item
  Events in this world are sometimes caused by divine intervention.
\item
  Some form of life continues after death.
\item
  God created the universe.
\item
  There is a heaven.
\end{enumerate}

The record diagnostics ask the following two questions after the log.

\begin{quote}
Over the twenty weeks in the log, how many weeks was Oren right, and how
many weeks was Teo right? Last line: ``Answer: Oren:
\textless number\textgreater, Teo: \textless number\textgreater{}''.
\end{quote}

\begin{quote}
If you had to ask one person in Harrow about this, Oren or Teo, who
would you ask? Last line: ``Answer: \textless name\textgreater{}''.
\end{quote}
\clearpage
\bibliography{references.bib}
\end{document}